\documentclass[11pt, a4paper, onecolumn, copyright, gdm]{google}

\usepackage[authoryear, sort&compress, round]{natbib}
\usepackage{caption}
\usepackage{subcaption}
\usepackage[capitalize,noabbrev]{cleveref}
\usepackage{cancel}
\usepackage{mathtools}
\usepackage{xcolor}
\usepackage{algorithm}
\usepackage{algpseudocode}
\usepackage{booktabs}
\usepackage{graphicx}
\usepackage{wrapfig}
\usepackage{pifont}

\def\N{\mathcal{N}}
\def\E{\mathbb{E}}

\def\I{\mathbf{I}}
\def\x{\mathbf{x}}
\def\z{\mathbf{z}}

\def\ee{\mathbf{e}}
\def\hx{\hat{\x}}
\def\tx{\tilde{\x}}

\def\a{\alpha}
\def\s{\sigma}
\def\d{\mathrm{d}}
\def\l{\lambda}
\def\e{\varepsilon}
\def\ph{\varphi}
\def\th{\theta}
\def\Th{\Theta}
\def\D{\Delta}

\def\n{\nabla}
\def\T{\mathrm{T}}
\def\KL{{\mathrm{D}_\mathrm{KL}}}
\def\L{\mathcal{L}}
\def\tL{\tilde{\L}}
\def\cmark{{\textcolor{green!50!black}{\ding{51}}}}
\def\xmark{{\textcolor{red!80!black}{\ding{55}}}}

\newcommand{\TODO}[1][]{{\ifthenelse{\isempty{#1}}{\color{red}\bf TODO}{\color{red}\bf (TODO: #1)}}}
\newcommand{\tm}[1]{{\ifthenelse{\isempty{#1}}{\color{cyan}\bf TM}{\color{cyan}\bf (TM: #1)}}}
\newcommand{\sg}[1]{\textrm{sg}\left( #1 \right)}

\renewcommand{\today}{} 
\title{Dual-Rate Diffusion: Accelerating diffusion models with an interleaved heavy-light network}
\correspondingauthor{bartosh@google.com, ruhe@google.com, mensink@google.com}

\author[*,1,2]{Grigory Bartosh}
\author[2]{David Ruhe}
\author[2]{Emiel Hoogeboom}
\author[2]{Jonathan Heek}
\author[2]{Thomas Mensink}
\author[2]{Tim Salimans}

\affil[*]{Work done while interning at Google DeepMind Amsterdam}
\affil[1]{University of Amsterdam}
\affil[2]{Google DeepMind Amsterdam}

\begin{abstract}
    Diffusion models achieve state-of-the-art generative performance but suffer from high computational costs during inference due to the repeated evaluation of a heavy neural network. In this work, we propose Dual-Rate Diffusion, a method to accelerate sampling by interleaving the execution of a heavy high-capacity context encoder and a light efficient denoising model. The context encoder is evaluated sparsely to extract high-dimensional features, which the light denoising model reuses at every step to refine the sample. This approach significantly accelerates inference without compromising sample quality. On ImageNet benchmarks, Dual-Rate Diffusion matches the performance of standard baselines while reducing computational cost by a factor of $2$--$4$. Furthermore, we demonstrate that our method is compatible with distillation techniques, such as Moment Matching Distillation, which enables further efficiency gains in few-step generation.
\end{abstract}

\uselogo{}  

\begin{document}
\maketitle

\section{Introduction}
\label{sec:introduction}

Diffusion models have recently demonstrated remarkable performance across various domains, including image \citep{dhariwal2021diffusion}, video \citep{ho2022video}, and audio generation \citep{kong2020diffwave}. Despite their impressive sample quality, these models are computationally expensive during inference, necessitating numerous function evaluations to generate a single sample.

To reduce inference time, recent research has proposed techniques such as distillation and direct training of few-step diffusion models. While these methods decrease the number of function evaluations, the underlying function typically remains parameterized by the same computationally heavy neural network at each step. At the same time, other studies \citep{chen2022analog, ma2024deepcache, liu2024faster} suggest that computations from earlier denoising steps can be efficiently reused, implying that a high-capacity network may not be necessary at every step.

\begin{figure}[t]
    \centering
    \includegraphics[width=\linewidth]{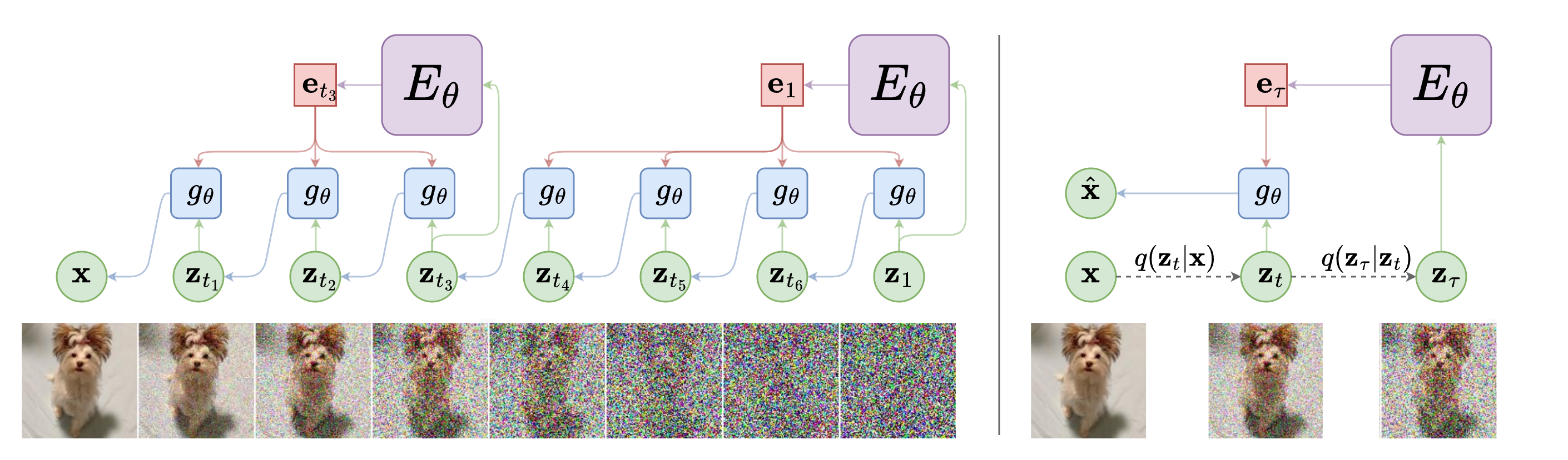}
    \caption{Overview of Dual-Rate Diffusion generative process (on the left) and training (on the right). The context encoder $E_\Th$ (heavy network) is evaluated at sparse intervals to extract high-dimensional features, which are then reused by the light denoising model $g_\th$ at every step to refine the sample.}
    \label{fig:method}
\end{figure}

In this work, we introduce Dual-Rate Diffusion, a simple technique to accelerate diffusion models by interleaving a heavy context encoder and a light denoising model, see \cref{fig:method}. The context encoder is evaluated at selected steps to compute high-dimensional features of the noisy samples. These features are then utilized by the denoising model to handle the denoising steps more efficiently.

This approach is motivated by two perspectives. First, it generalizes self-conditioning \citep{chen2022analog}, where the model's own prediction of the clean data from the previous step is fed back as an additional input: in Dual-Rate Diffusion, the conditioning is not restricted to the form or dimensionality of the data, and it is produced by a separate network, which decouples the per-step computation from the network that extracts the global structure. Second, we can view the denoising process in the frequency domain, where the global structure (low frequencies) is established in early steps and remains stable, while local details emerge later \citep{rissanen2023generative, dieleman2024spectral}. Evaluating the denoising model independently at every time step leads to repeated, redundant processing of this global structure, which consumes the model capacity. Dual-Rate Diffusion delegates this task to the heavy context encoder, evaluated once per interval of steps; in the steps that follow, the light denoising model reuses the extracted information and focuses on local structure.

We demonstrate that Dual-Rate Diffusion significantly reduces inference time for standard diffusion models without compromising sample quality. On ImageNet benchmarks, Dual-Rate Diffusion matches the sample quality of standard baselines while reducing the computational cost by a factor of $2$--$4$. Furthermore, we show that this method is compatible with distillation techniques such as MMD \citep{salimans2024multistep}, where Dual-Rate students sample at a lower cost than standard MMD students and improve on them on ImageNet $64\times64$.

\section{Background}
\label{sec:background}

\textbf{Diffusion Models} ~
Diffusion models \citep{ho2020denoising} consist of a forward process and a reverse process. The forward process gradually corrupts a data point $\x$ by injecting Gaussian noise. Intermediate noisy states $\z_t \sim q(\z_t|\x)$ at time step $t \in [0, 1]$ can be generated as:
\begin{align}
    \z_t = \a_t \x + \s_t \e, \quad \text{where} \quad \e \sim \N(0, \I).
\end{align}
We employ a variance-preserving noise schedule \citep{song2021scorebased}, where $\a_t^2 + \s_t^2 = 1$, defined by the log signal-to-noise ratio (log-SNR) $\l_t = \log (\a_t^2 / \s_t^2)$, which is a monotonic function of $t$.

The reverse process is an iterative refinement process. It starts from pure noise $\z_1$ and gradually denoises it to recover a data sample $\z_0$. At each iteration, the model predicts the clean data $\hx = g_\th(\z_t, t)$ and then samples a less noisy state $\z_s$ (with $s<t$) from the posterior distribution $q(\z_s|\z_t, \x=\hx)$. Finally, $\x$ is set to $\z_0$ exactly or quantized.

As the number of steps approaches infinity, a variational bound on the model log-likelihood is given by \citet{ho2020denoising}:
\begin{align}
    \KL \big( q(\x) \| p_\th(\x) \big)
    &\leq \KL \big( q(\x, \z_0, \dots, \z_1) \| p_\th(\x, \z_0, \dots, \z_1) \big) \nonumber \\
    &\approx \tfrac{1}{2} \E_{u(t)q(\x,\z_t)} \Big[ -\tfrac{\d \l_t}{\d t} e^{\l_t} w(\l_t) \big\| \x - g_\th(\z_t, t) \big\|^2 \Big] + \underbrace{\KL \big( q(\z_1) \| p(\z_1) \big)}_{\approx 0},
    \label{eq:elbo}
\end{align}
where $u(t)$ is a uniform distribution over $t \in [0, 1]$ and $w(\l_t)$ is a weighting function. The objective is minimized when $g_\th(\z_t, t) = \E_q[\x|\z_t]$. The bound holds strictly for $w(\l_t) \equiv 1$. In practice, however, a re-weighted ELBO (e.g., $w(\l_t) = \text{sigmoid}(\l_t - b)$) is often used to improve sample quality \citep{kingma2023understanding}.

\textbf{Few-Step Generation and Moment Matching Distillation} ~
Few-step generative models are commonly obtained by distilling a pretrained diffusion model, and existing methods fall into two broad families. Trajectory-based methods, such as progressive distillation \citep{salimans2022progressive} and consistency models \citep{song2023consistency}, train the student to reproduce the deterministic sampling trajectory of the teacher in fewer steps. Distribution matching methods, such as DMD \citep{yin2023one} and Moment Matching Distillation (MMD) \citep{salimans2024multistep}, instead train the student so that the distribution of its samples matches the distribution induced by the teacher. In this work, we focus on MMD.

MMD adapts the standard iterative refinement process for few-step generation. Given a noisy sample $\z_t$, rather than estimating the expectation $\E_q[\x|\z_t]$, MMD generates samples $\tx$ from a learnable distribution $p_\eta(\tx|\z_t)$ and subsequently samples $\z_s$ from the forward posterior $q(\z_s|\z_t, \x=\tx)$. This procedure implicitly defines the distribution $p_\eta(\tx|\z_s)$. It can be shown that if $p_\eta(\tx|\z_s) \equiv q(\x|\z_s)$, the resulting process yields correct marginals $p_\eta(\z_s) = q(\z_s)$ for any $s<t$. Matching these conditional distributions is thus the proper objective; MMD approximates it by matching their first moments, minimizing the $\mathrm{L}2$-distance:
\begin{align}
    \tL = \frac{1}{2} \E_{q(\z_t) p_\eta(\z_s|\z_t)} \left[ \left\| \E_{p_\eta}[\tx|\z_s] - \E_q[\x|\z_s] \right\|^2  \right].
    \label{eq:mmd_objective}
\end{align}
This objective can be reformulated to simplify optimization while preserving gradients $\n_\eta \tL = \n_\eta \L$:
\begin{align}
    \L = \E_{q(\z_t) p_\eta(\tx, \z_s|\z_t)} \Big[ \tx^\T \sg{ \E_{p_\eta}[\tx|\z_s] - \E_q[\x|\z_s] } \Big],
\end{align}
where $\sg{\cdot}$ denotes the stop-gradient operator. The intractable moments in this objective are approximated using an auxiliary model ${g_\ph(\z_t, t) \approx \E_{p_\eta}[\tx|\z_s]}$ and a teacher model ${g_\th(\z_t, t) \approx \E_q[\x|\z_s]}$. Both $g_\ph$ and $g_\th$ can be trained via the standard diffusion loss~(\cref{eq:elbo}), with $g_\th$ potentially being a pretrained diffusion model. In practice, MMD uses a deterministic student model $\tx = g_\eta(\z_t, t)$ which learns to generate samples matching the teacher model's moments.

\section{Dual-Rate Diffusion}
\label{sec:method}

Dual-Rate Diffusion accelerates denoising by employing a dual-network architecture: a computationally heavy network (context encoder) and a light network (denoising model). The context encoder is evaluated only at sparse, fixed intervals (e.g., $16$ times along a $512$-step sampling trajectory) and extracts high-dimensional features that capture the global structure of the sample. The denoising model runs at every step and reuses the most recent features instead of recomputing the global structure, which lets it focus its computation on refining local details~(see \cref{fig:method}). In this section, we present the sampling and training procedures of Dual-Rate Diffusion in the standard diffusion setting and then show how the construction applies to distillation, using MMD as an example.

\subsection{Diffusion}
\label{sec:method_diffusion}

The context encoder processes a noisy sample $\z_\tau$ at a specific time step $\tau$ to generate a high-dimensional feature vector $\ee_\tau = E_\Th(\z_\tau, \tau)$. The denoising model then predicts the clean data $\hx = g_\th(\z_t, t, \ee_\tau)$ using the current noisy sample $\z_t$, the current time step $t \le \tau$, and the most recently computed feature $\ee_\tau$, where $\tau$ refers to the closest preceding evaluation step of the context encoder. If the denoising model is evaluated $k$ times during sampling, the context encoder is evaluated $K \leq k$ times, once every $k/K$ steps of the denoising model, where $K$ is a divisor of $k$. Whenever the context encoder is evaluated, it is evaluated on the current state, so $\z_\tau = \z_t$ at that step, and the features $\ee_\tau$ are then held fixed for the following $k/K - 1$ denoising steps. Each denoising step is otherwise a standard ancestral step: the denoising model predicts the clean data, and the next state is sampled from the forward posterior $q(\z_s|\z_t, \hx)$. The cost of a trajectory is therefore $K$ evaluations of the heavy network plus $k$ evaluations of the light one, and the ratio $k/K$ sets the trade-off between the two. This sampling procedure is summarized in \cref{alg:dual_rate_sampling}.

\begin{algorithm}[t]
\caption{Ancestral sampling algorithm for Dual-Rate Diffusion.}
\label{alg:dual_rate_sampling}
\begin{algorithmic}[1]
\Require Context encoder $E_\Th$, denoising model $g_\th$, number of context encoder evaluations $K$ and denoising steps $k$.
\State Initialize noisy data $\z_1 \sim \N(0,\I)$
\For{$t \in \{1, (k-1)/k, \dots, 1/k$\}}
    \If{$t \in \{1, (K-1)/K, \dots, 1/K\}$}
        \State Evaluate context encoder $\ee_\tau = E_\Th(\z_t, t)$
    \EndIf
    \State Predict clean data with denoising model $\hx = g_\th(\z_t, t, \ee_\tau)$
    \State Set next time step $s = t - 1/k$
    \State Sample next noisy point $\z_s \sim q(\z_s | \z_t, \hx)$
\EndFor
\State \Return Approximate sample $\hx$
\end{algorithmic}
\end{algorithm}

\begin{figure}[t]
    \centering
    \includegraphics[width=\linewidth]{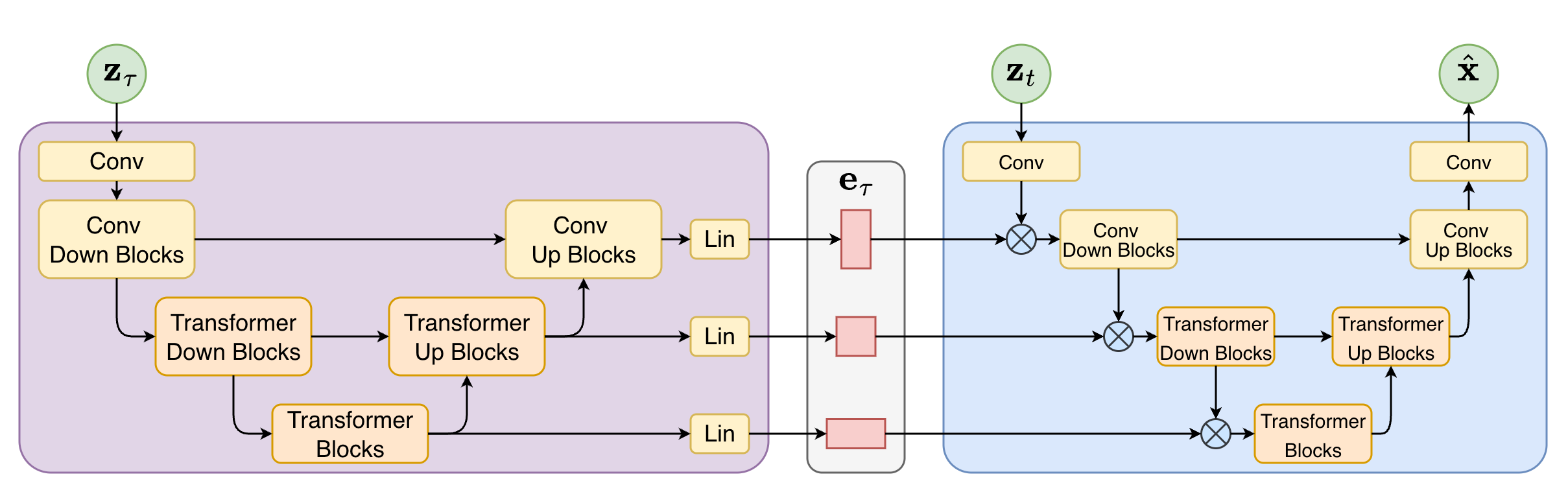}
    \caption{Overview of the Dual-Rate Diffusion conditioning mechanism. The context encoder (on the left) generates high-dimensional features at each level of the UViT architecture, which are then reused at the corresponding level of the denoising UViT (on the right). The $\otimes$ symbol denotes concatenation.}
    \label{fig:conditioning}
\end{figure}

\textbf{Conditioning mechanism.} We parameterize both networks with hierarchical UViT architectures \citep{hoogeboom2025simpler}; \cref{fig:conditioning} depicts how the two are connected. The features $\ee_\tau$ are the outputs of the last layer at each spatial resolution level of the context encoder, so they retain the spatial layout of the sample at every level rather than being pooled into a single vector. At the beginning of the corresponding level of the denoising model, they are linearly projected to match the dimensions and concatenated with the denoising model's own features, which lets each level of the light network start from the representation the heavy network computed at the same level. The denoising model is additionally conditioned on the time step $\tau$, which tells it how stale the features are; we omit this from the notation $g_\th(\z_t, t, \ee_\tau)$.

\textbf{Training.} The Dual-Rate Diffusion objective extends the standard diffusion loss~(\cref{eq:elbo}):
\begin{align}
    \KL \big( q(\x) \| p_{\Th, \th}(\x) \big)
    &\leq \frac{1}{2} \E_{u(t, \tau)q(\x,\z_t,\z_\tau)} \left[ -\frac{\d \l_t}{\d t} e^{\l_t} w(\l_t) \left\| \x - g_\th \big( \z_t, t, E_\Th(\z_\tau, \tau) \big) \right\|^2 \right],
\end{align}
where $u(t, \tau)$ denotes a uniform distribution over the joint time steps. In our experiments, we sample $\tau$ uniformly from the interval $[1/K, 1]$ and subsequently sample $t$ uniformly from $(\tau - 1/K, \tau]$. This objective remains a valid variational bound on the model likelihood when $w(\l_t)=1$. \cref{alg:dual_rate_training} details the training procedure.

\begin{algorithm}[t]
\caption{Dual-Rate Diffusion training algorithm.}
\label{alg:dual_rate_training}
\begin{algorithmic}[1]
\Require Context encoder $E_\Th$, denoising model $g_\th$, number of context encoder evaluations $K$ and denoising steps $k$, loss weight $w(\l_t)$, and dataset~$\mathcal{D}$.
\For{$n=0$:$N$}
    \State Sample clean data $\x \sim \mathcal{D}$
    \State Sample time of context encoder evaluation $\tau \sim u[1/K, 1]$
    \State Sample time delta $\delta_t \sim u(0, 1/K]$ and set $t = \tau - \delta_t$
    \State Sample noisy state $\z_\tau \sim q(\z_\tau | \x)$ and less noisy $\z_t \sim q(\z_t | \z_\tau, \x)$
    \State Minimize $\L = -\frac{\d \l_t}{\d t} e^{\l_t} w(\l_t) \left\| \x - g_\th \big( \z_t, t, E_\Th(\z_\tau, \tau) \big) \right\|^2$ with respect to $\th$ and $\Th$
\EndFor
\end{algorithmic}
\end{algorithm}

\textbf{Markovian processes.} Unlike standard diffusion, correct sampling in Dual-Rate Diffusion generally requires matching the joint distribution $p_{\Th, \th}(\z_t, \z_\tau) \approx q(\z_t, \z_\tau)$ during inference. Since the denoising model $g_\th$ depends on both $\z_t$ and $\z_\tau$, using non-Markovian posteriors $q(\z_s|\x,\z_t)$ (e.g., DDIM \citep{song2020denoising}) causes a distribution mismatch between training and inference. Training with the same non-Markovian posterior does not resolve this. With a Markovian posterior, the generative transitions $p(\z_s|\z_t)$ approach the forward transitions $q(\z_s|\z_t)$ as the model approaches the data posterior, and the joint distribution of $(\z_t, \z_\tau)$ is determined by these transitions, exactly as in standard diffusion. With a non-Markovian posterior, every state between $\z_\tau$ and $\z_t$ is tied to the same clean sample $\x$ under the forward process, whereas the generative process replaces $\x$ by a new prediction $\hx$ at every step, so the two induce different joint distributions of $(\z_t, \z_\tau)$ regardless of how the model is trained. Therefore, we use Markovian processes for both training and sampling; the effect of applying a deterministic sampler to a trained model is quantified in \cref{sec:experiments_diffusion}.

The Markovian requirement concerns the target process, not the model's own. Each generative step depends on two previous states, $\z_t$ and $\z_\tau$, so the process the model defines is itself non-Markovian; its target is Markovian: $\z_\tau$ carries no information about the next state beyond what $\z_t$ already contains, and a sufficiently flexible denoising model alone would suffice. The context encoder is therefore a computational device rather than a modelling one: it extracts, once per interval, the information that a light denoising model cannot afford to recompute at every step, and packages it for reuse. The combined model thus spends its capacity and compute where they are needed while remaining as flexible as a larger denoising model alone.

\subsection{Distillation}
\label{sec:method_distillation}

\begin{algorithm}[t]
\caption{Dual-Rate MMD training algorithm with alternating optimization of auxiliary denoising model $g_\ph$ and student models $E_H$ and $g_\eta$. Differences from standard MMD are highlighted in \textcolor{blue}{blue}.}
\label{alg:dual_rate_mmd_training}
\begin{algorithmic}[1]
\Require Pretrained denoising model $g_\th$, student context encoder $E_H$ and denoising model $g_\eta$, auxiliary denoising model $g_\ph$, number of context encoder evaluations $K$ and denoising steps $k$, loss weight $w(\l_t)$, dataset~$\mathcal{D}$.
\For{$n=0$:$N$}
    \State Sample clean data $\x \sim \mathcal{D}$
    \State \textcolor{blue}{Sample time of context encoder evaluation $\tau \sim U(\{1/K, \dots, (K-1)/K, 1\})$}
    \State \textcolor{blue}{Sample step delta $\D_t \sim U(\{0, 1, \dots, k/K - 1\})$ and set $t = \tau - \frac{\D_t}{k}$}
    \State Sample time delta $\delta_s \sim u(0, 1/k]$ and set $s = t - \delta_s$
    \State \textcolor{blue}{Sample noisy state $\z_\tau \sim q(\z_\tau | \x)$ and calculate $\ee_\tau = E_H(\z_\tau, \tau)$}
    \State \textcolor{blue}{Sample $\z_t \sim p_{H,\eta}(\z_t | \z_\tau)$ with a rollout of the denoising model $g_\eta$}
    \State Sample $\z_s$ as $\tx = g_\eta(\z_t, t, \textcolor{blue}{\ee_\tau}), \z_s \sim q(\z_s | \z_t, \tx)$
    \If{$n$ is even}
        \State Minimize $\L = -\frac{\d \l_s}{\d s} e^{\l_s} w(\l_s) \left( \left\| \tx - g_\ph(\z_s, s) \right\|^2 + \left\| g_\th(\z_s, s) - g_\ph(\z_s, s) \right\|^2 \right)$ w.r.t. $\ph$
    \Else
        \State Minimize $\L = -\frac{\d \l_s}{\d s} e^{\l_s} w(\l_s) \tx^\T \sg{g_\ph(\z_s, s) - g_\th(\z_s, s)}$ w.r.t. $\eta$ and $H$
    \EndIf
\EndFor
\end{algorithmic}
\end{algorithm}

Although distillation and direct training can yield one-step models, achieving high sample quality often requires maintaining a multi-step process. However, even for these few-step models, evaluating a deep neural network at every step can be computationally prohibitive. Dual-Rate Diffusion addresses this bottleneck by amortizing the cost of heavy global computations across multiple lighter steps, replacing the heavy network with the light denoising model for the intermediate refinements.

A generic approach would be to take any model (distilled or not), sample generative trajectories, and perform a secondary distillation step by training a Dual-Rate model on these trajectories. While valid, this strategy complicates the training pipeline and may introduce compounding errors from the additional distillation stage. Instead, we demonstrate that Dual-Rate Diffusion can be trained directly as a few-step model within Moment Matching Distillation (MMD), without this extra stage.

To construct a Dual-Rate MMD model, we simply substitute the Dual-Rate architecture for the student model of MMD. We denote the student context encoder as $E_H$ and the student denoising model as $g_\eta$, with $H$ and $\eta$ their parameters; the teacher $g_\th$ and the auxiliary model $g_\ph$ remain standard single-network denoising models as in \cref{sec:background}. For sampling, we employ the exact same procedure as in standard Dual-Rate Diffusion (see \cref{alg:dual_rate_sampling}).

The Dual-Rate MMD objective directly extends the standard MMD loss~(\cref{eq:mmd_objective}) by sampling not two but three noisy states at $s < t \leq \tau$:
\begin{align}
    \L = \E_{q(\z_\tau) p_{H,\eta}(\tx, \z_s, \z_t|\z_\tau)} \Big[ \tx^\T \sg{ \E_{p_{H,\eta}}[\tx|\z_s] - \E_q[\x|\z_s] } \Big].
\end{align}

Similar to regular MMD, training requires a pretrained denoising teacher model $g_\th$ and an auxiliary denoising model $g_\ph$ to approximate the intractable moments $\E_q[\x|\z_s]$ and $\E_{p_{H,\eta}}[\tx|\z_s]$, respectively.

While the MMD objective targets marginal distribution matching, $p_{H,\eta}(\z_s) \approx q(\z_s)$, it does not guarantee that the student's generated trajectory distribution matches the ground truth forward process. Consequently, if the student is trained using intermediate states sampled from the true forward process $q(\z_t, \z_\tau)$, it will encounter out-of-distribution inputs during inference, leading to compounding errors.

To ensure that the student stays in-distribution during training, intermediate states $\z_t$ must be generated by the model itself, mirroring the inference dynamics. Specifically, rather than relying on the teacher's forward process, we sample $\z_t \sim p_{H,\eta}(\z_t | \z_\tau)$ by unrolling the student's iterative sampling procedure. Thus, whenever $t$ and $\tau$ span multiple denoising steps, we perform a sequence of generative steps using the denoising model $g_\eta$ to traverse from $\z_\tau$ to $\z_t$. We refer to this process as a rollout of the denoising model $g_\eta$. The complete training procedure is summarized in \cref{alg:dual_rate_mmd_training}. The same principle extends to $\z_\tau$: instead of sampling it from the forward process (line 6), the student can generate it as well, by a full rollout from pure noise. This aligns the input distribution of the student during training even more closely with inference, and since every state is then generated by the student, distillation no longer requires training data; all reported results use this variant (\cref{app:ablations_mmd}).

Following the original MMD setup, we alternate between optimizing the auxiliary model $g_\ph$ and the student models $E_H$ and $g_\eta$. The auxiliary model is trained to predict the samples $\tx$ while being regularized to remain close to the teacher model $g_\th$. This regularization helps stabilize the training process and should not bias the optimum point.

In summary, Dual-Rate MMD differs from the original MMD formulation in three key aspects. First, it parameterizes the student model with a Dual-Rate architecture. Second, intermediate states $\z_t$ are generated via a rollout of the student's denoising model to prevent distribution shift. Finally, whereas the original MMD samples the time steps $t$ from a continuous uniform distribution, we sample both $t$ and $\tau$ from discrete uniform distributions defined over the inference grids of denoising steps and context encoder evaluations. This follows the same in-distribution principle as the rollout: at inference, the student is evaluated only on the fixed grid of $k$ denoising steps and $K$ context encoder evaluations (\cref{alg:dual_rate_sampling}), and sampling $t$ and $\tau$ continuously exposes it to time pairs that never occur at inference, which in our experiments destabilized the early phase of distillation. The time $s$, at which only the teacher and the auxiliary model are evaluated, remains continuous.

\textbf{Other distillation methods.} The construction above is not specific to moment matching: the three modifications are defined for any objective evaluated on sampled states of a stochastic few-step trajectory, such as other distribution matching methods \citep{yin2023one}; we test only MMD. Objectives built on a deterministic teacher trajectory, such as progressive distillation \citep{salimans2022progressive} and consistency models \citep{song2023consistency}, or on a differential characterization of the flow map \citep{boffi2025how}, do not satisfy the Markovian requirement of \cref{sec:method_diffusion}; models trained with them can be converted into Dual-Rate models by the trajectory distillation described above.

\section{Experiments}
\label{sec:experiments}

We evaluate Dual-Rate Diffusion on the standard class-conditional ImageNet \citep{deng2009imagenet} benchmarks at $64\times64$ and $128\times128$ in two settings: (1) training a standard diffusion model to demonstrate the efficiency gains of the interleaving architecture, and (2) applying Dual-Rate Diffusion to Moment Matching Distillation (MMD) to show its compatibility with few-step distillation methods.

\textbf{Metrics.} In all experiments, we report the Fréchet Inception Distance (FID) \citep{heusel2017gans} as the primary metric for sample quality, computed on 50k generated samples against the training set. We measure the computational cost of sampling in FLOPs, which are hardware-independent, and also report the number of function evaluations (NFE).

\textbf{Baselines.} We compare against pixel-space diffusion models on the same benchmarks. VDM\texttt{++} \citep{kingma2023understanding} and EDM2 \citep{karras2024analyzing} are standard diffusion models with many-step samplers, the latter using a deterministic second-order Heun solver; SiD2 \citep{hoogeboom2025simpler} is the model whose training recipe our base model follows. In the few-step setting, CTM \citep{kim2023consistency} and sCD \citep{lu2024simplifying} are consistency-type models trained along the deterministic trajectory of a teacher, DMD2 \citep{yin2024improved} is a distribution matching method, and Moment Matching Distillation \citep{salimans2024multistep} is the distribution matching method of \cref{sec:background} on which our distilled models build.

\textbf{Experimental setup.} For the regular diffusion experiments, we follow the setup from \citet{hoogeboom2025simpler}, including sigmoid weighting of the objective and an optimized amount of sampling noise, following \citet{salimans2022progressive}. For the distillation experiments, we follow the original MMD setup from \citet{salimans2024multistep}, using a pretrained standard diffusion model as the teacher. We parameterize the context encoder $E_\Th$ with the UViT architecture of the base model, and the denoising model $g_\th$ with the same architecture with fewer blocks and channels, approximately two times lighter in FLOPs; the exact architectures are given in \cref{tab:implementation}, further details in \cref{app:implementation}, and sample outputs in \cref{app:samples}.

\subsection{Standard Diffusion}
\label{sec:experiments_diffusion}

We first validate Dual-Rate Diffusion in the standard diffusion setting.

\begin{figure}[t]
    \centering
    \begin{minipage}[t]{0.48\textwidth}
        \vspace{0pt}
        \centering
        \captionof{table}{Ablation of different training techniques for Dual-Rate Diffusion on ImageNet $64\times64$ with $K=16$ context encoder evaluations and $k=512$ denoising steps.}
        \label{tab:i64_ablations}
        \vspace{6pt}
        \small
        \begin{tabular}[t]{cccc}
            \toprule
            Multi-level & Data Aug & Feature Drop & FID \\
            \midrule
            \xmark & \xmark & \xmark & 1.30 \\
            \cmark & \xmark & \xmark & 1.24 \\
            \xmark & \cmark & \xmark & 1.26 \\
            \xmark & \xmark & \cmark & 1.14 \\
            \cmark & \cmark & \xmark & 1.20 \\
            \cmark & \xmark & \cmark & 1.13 \\
            \cmark & \cmark & \cmark & \textbf{1.12} \\
            \bottomrule
        \end{tabular}
    \end{minipage}
    \hfill
    \begin{minipage}[t]{0.48\textwidth}
        \vspace{0pt}
        \centering
        \includegraphics[width=\linewidth]{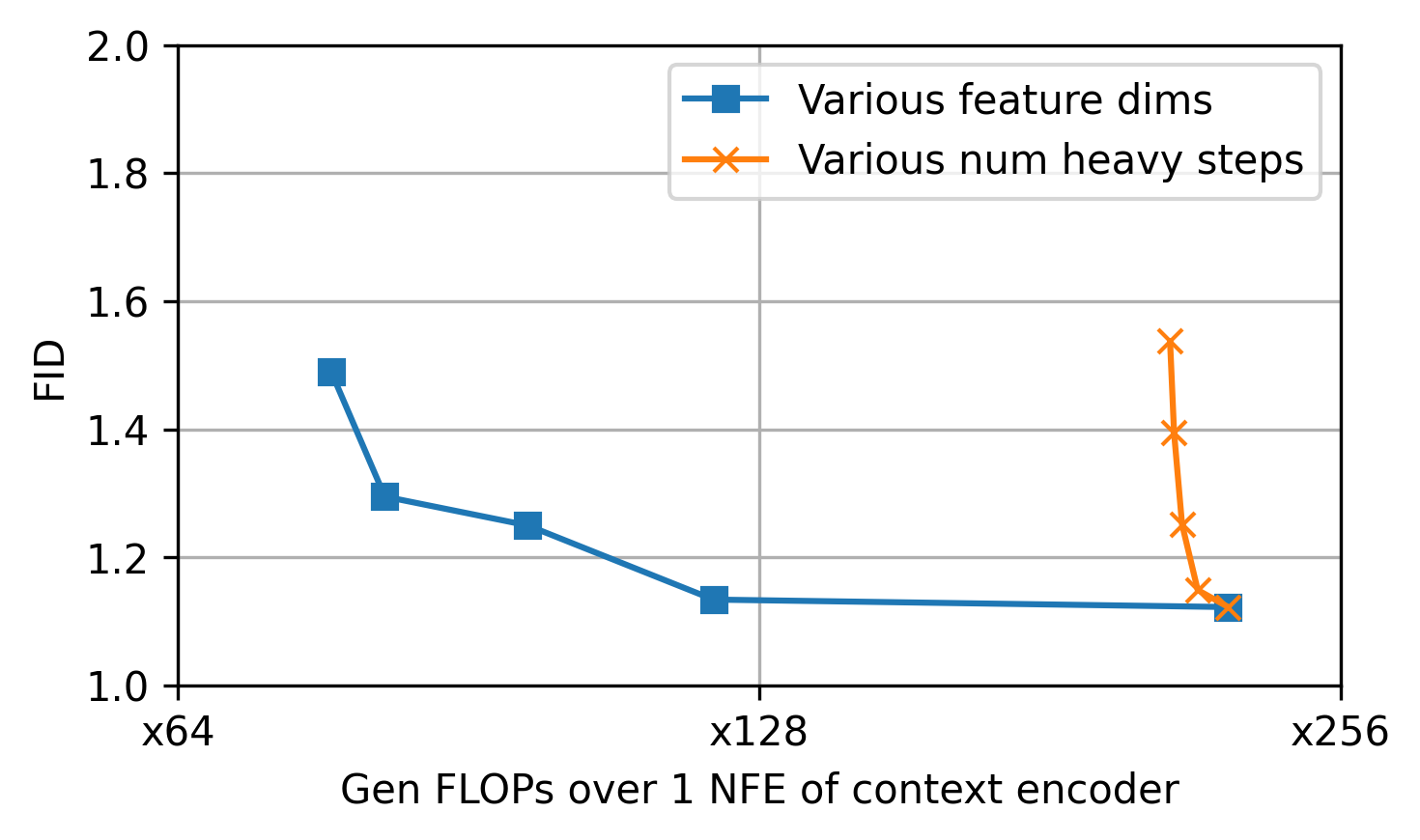}
        \caption{Ablation of the number of context encoder evaluations $K$ versus the feature dimension of the denoising model (ImageNet $64\times64$).}
        \label{fig:i64_features_vs_big_steps}
    \end{minipage}
\end{figure}

Three training techniques improve Dual-Rate Diffusion. First, we condition the denoising model on features from every level of the context encoder hierarchy (\cref{sec:method_diffusion}) rather than only the last level. Second, we use extra data augmentation (random translation) during training. Third, during training we randomly replace the features $\ee_\tau$ with a zero vector with probability $p=0.5$. This feature dropout mirrors the training procedure of self-conditioning \citep{chen2022analog}; our hypothesis is that, because the denoising model must also work without them, the context features it learns to use complement its own computation rather than replace it, which makes them more robust. \cref{tab:i64_ablations} ablates the three techniques: all of them contribute to the performance improvement, and feature dropout has the largest effect on the FID score.

\begin{wrapfigure}{r}{0.5\textwidth}
    \vspace{-14pt}
    \centering
    \includegraphics[width=\linewidth]{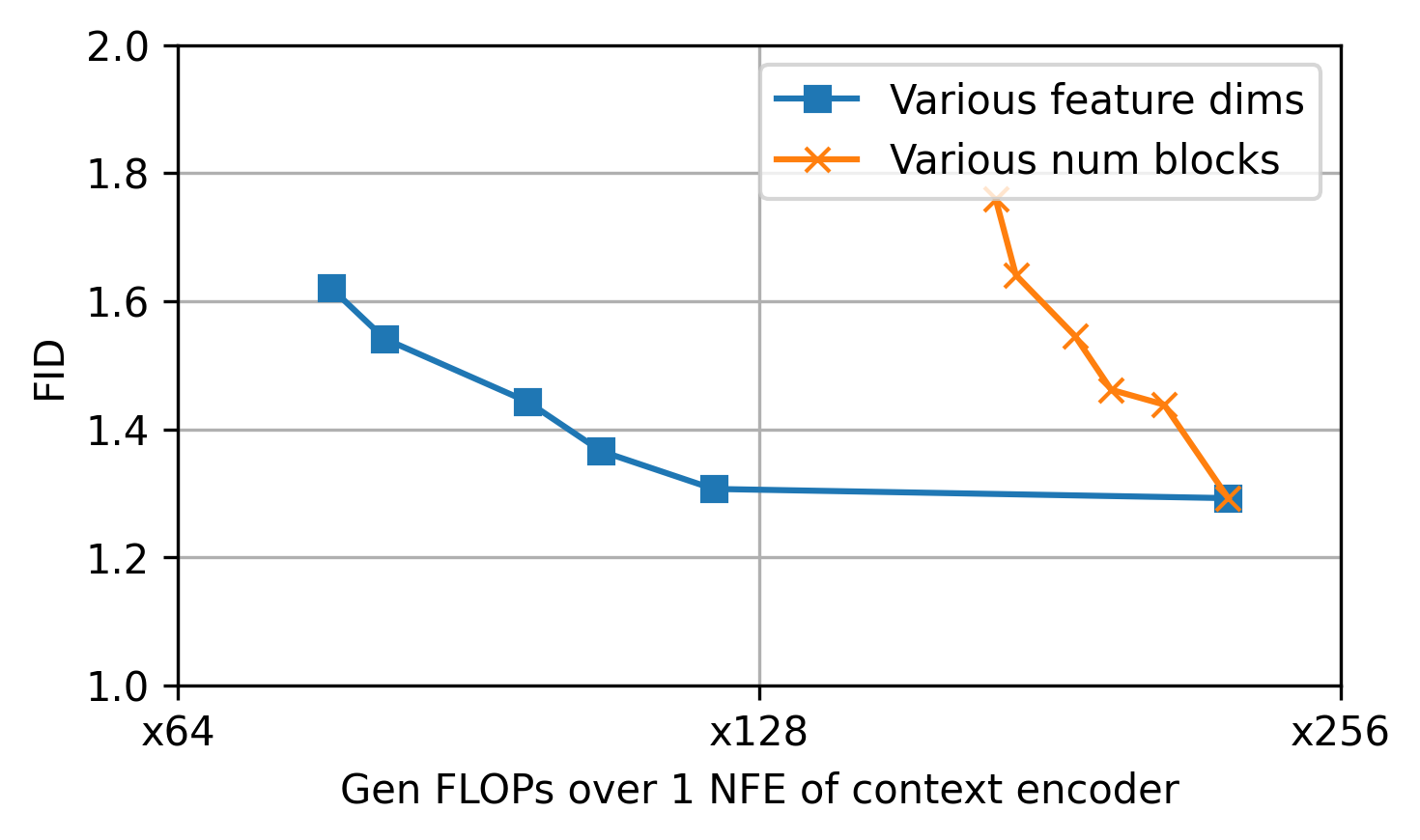}
    \caption{Ablation of the denoising model depth versus its feature dimension on ImageNet $64\times64$.}
    \label{fig:i64_features_vs_blocks}
    \vspace{-10pt}
\end{wrapfigure}

We also ablate the number of context encoder evaluations and the capacity allocation within the denoising model. \cref{fig:i64_features_vs_big_steps} compares reducing $K$ with narrowing the denoising model ($512$ down to $128$ channels per level): reducing $K$ gives a worse performance-compute trade-off, and the FID degrades sharply below $8$ evaluations. \cref{fig:i64_features_vs_blocks} ablates depth against width: removing blocks hurts markedly, whereas narrowing offers a much better trade-off. \cref{app:ablations_diffusion} gives the exact ranges of these ablations and a comparison with fully convolutional denoising models, which performed considerably worse.

\begin{table}[t]
\centering
\caption{Results on ImageNet $64\times 64$ and $128\times 128$ for Dual-Rate Diffusion and Dual-Rate MMD. We report the computational cost of sampling in terms of teraflops (TFLOPs) and number of function evaluations (NFE). For the Dual-Rate experiments NFE refers to the number of denoising steps and $K$ refers to the number of context encoder evaluations. The Moment Matching results are those reported by \citet{salimans2024multistep} and were obtained with their teacher model (see \cref{sec:experiments_mmd}).}
\label{tab:results}
\resizebox{.96\textwidth}{!}{
\begin{tabular}{lccccc}
\toprule
 & & \multicolumn{2}{c}{\textbf{ImageNet $64\times 64$}} & \multicolumn{2}{c}{\textbf{ImageNet $128\times 128$}} \\
Methods & NFE & TFLOPs & FID$\downarrow$ & TFLOPs & FID$\downarrow$ \\
\midrule
\multicolumn{6}{c}{\sc{Base models}}\\
VDM$++$ \citep{kingma2023understanding} & $1024$ & $429.25$ & $1.43$ & $544.26$ & $1.75$ \\
EDM2-XL (Heun) \citep{karras2024analyzing} & $63$ & $25.58$ & $1.33$ & - & - \\
SiD2 \citep{hoogeboom2025simpler} & $512$ & - & - & $70.37$ & $1.26$ \\
Our base model (guidance $w=1$) & $512$ & $55.53$ & $1.25$ & $70.37$ & $1.50$ \\
Our base model (guidance $w=0$) & $512$ & $55.53$ & $2.38$ & $70.37$ & $2.66$ \\
\cmidrule(lr){1-6}
\textbf{Dual-Rate Diffusion ($K=16$)} \\
 & $128$ & $3.53$ & $1.52$ & $8.05$ & $3.21$ \\
 & $256$ & $7.44$ & $\textbf{1.12}$ & $13.98$ & $1.89$ \\
 & $512$ & $13.14$ & $\textbf{1.12}$ & $28.23$ & $\textbf{1.48}$ \\
\midrule
\multicolumn{6}{c}{\sc{Distillation}}\\
CTM \citep{kim2023consistency}
 & $1$ & - & $1.73$ & - & - \\
sCD-XXL \citep{lu2024simplifying}
 & $1$ & - & $2.04$ & - & - \\
 & $2$ & - & $1.48$ & - & - \\
Moment Matching \citep{salimans2024multistep}
 & $2$ & $0.22$ & $3.86$ & $0.27$ & $3.14$ \\
 & $4$ & $0.43$ & $1.50$ & $0.55$ & $1.72$ \\
 & $8$ & $0.87$ & $1.24$ & $1.10$ & $1.49$ \\
DMD2 \citep{yin2024improved}
 & $1$ & - & $1.28$ & - & - \\
\cmidrule(lr){1-6}
\textbf{Dual-Rate MMD} \\
\quad\quad\quad\quad$K=1$ & $2$ & $0.20$ & $1.56$ & $0.24$ & $2.60$ \\
\quad\quad\quad\quad$K=2$ & $4$ & $0.39$ & $1.44$ & $0.48$ & $2.58$ \\
\quad\quad\quad\quad$K=4$ & $8$ & $0.79$ & $1.17$ & $0.92$ & $1.90$ \\
\bottomrule
\end{tabular}
}
\end{table}

We summarize the evaluation results and compare with other pixel-space generative baselines in \cref{tab:results}. With a significantly lower sampling budget, Dual-Rate Diffusion achieves FID scores comparable to the base model at both resolutions. On ImageNet $64\times64$, it achieves a better FID score than the base model at the same number of denoising steps with a $4$ times smaller budget, and reaches the same quality with half the denoising steps at more than $7$ times lower cost.

\textbf{Capacity and compute.} The Dual-Rate model consists of the context encoder, which has the architecture of the base model, and the denoising model on top of it, so it has more parameters than the base model (\cref{tab:implementation}) but spends fewer FLOPs per sample. We attribute the improvement over the base model on ImageNet $64\times64$ to this combination: a higher effective capacity, used more efficiently during sampling. A single network with the per-step cost of our denoising model does not give the same trade-off. \cref{fig:i64_cw_vs_base} compares the base model, this lighter single network, and Dual-Rate Diffusion over a range of denoising steps: the lighter network is worse than Dual-Rate Diffusion at every budget, and falls behind the base model once the budget allows the base model enough steps.

\textbf{Training cost.} The efficiency gains of Dual-Rate Diffusion are realized at inference; training is more expensive than for the base model, because every training iteration evaluates both the context encoder and the denoising model (\cref{alg:dual_rate_training}). Each network is evaluated once per iteration with the same forward-backward structure as the base model, so the overhead per iteration is the ratio of the summed per-network FLOPs to those of the base model: about $1.4$ times on ImageNet $64\times64$ and $1.5$ times on ImageNet $128\times128$ (\cref{tab:training_cost}). Both models are trained for the same number of iterations with the same batch size, so the same factor applies to the total training compute. \cref{tab:training_cost} also compares the two models at matched training compute, by evaluating Dual-Rate Diffusion at the iteration at which it has consumed the total training compute of the fully trained base model. At that point it surpasses the base model on ImageNet $64\times64$ and is slightly behind it on ImageNet $128\times128$, while sampling $2$--$4$ times cheaper.

\textbf{Deterministic sampling.} Dual-Rate Diffusion is trained for the Markovian ancestral sampler, and a deterministic sampler such as DDIM \citep{song2020denoising} changes the joint distribution of $\z_t$ and $\z_\tau$ that the denoising model is evaluated on (\cref{sec:method_diffusion}). To measure its effect, we sampled the trained models with an untuned DDIM sampler on the same grid of $K=16$ context encoder evaluations and $k=512$ denoising steps. The mismatch costs $0.2$--$0.3$ FID at both resolutions (\cref{tab:ddim}); deterministic sampling therefore remains usable, and the ancestral sampler is our default.

\subsection{Distillation}
\label{sec:experiments_mmd}

\begin{wrapfigure}{r}{0.43\textwidth}
    \vspace{-14pt}
    \centering
    \includegraphics[width=\linewidth]{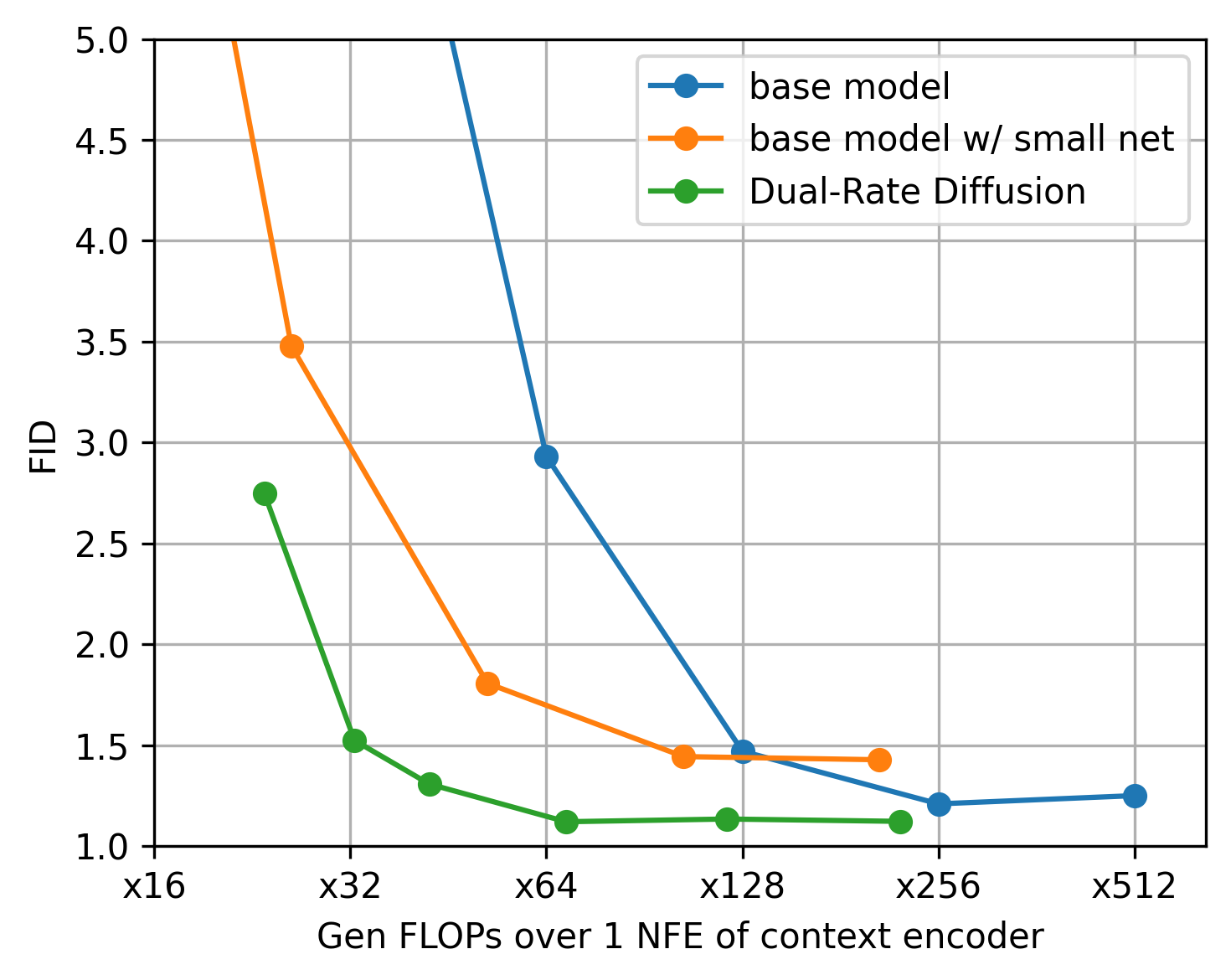}
    \caption{FID versus sampling cost on ImageNet $64\times64$ for the base model, a lighter single network, and Dual-Rate Diffusion. Cost in multiples of one context encoder evaluation ($108.45$ GFLOPs).}
    \label{fig:i64_cw_vs_base}
    \vspace{8pt}
    \includegraphics[width=\linewidth]{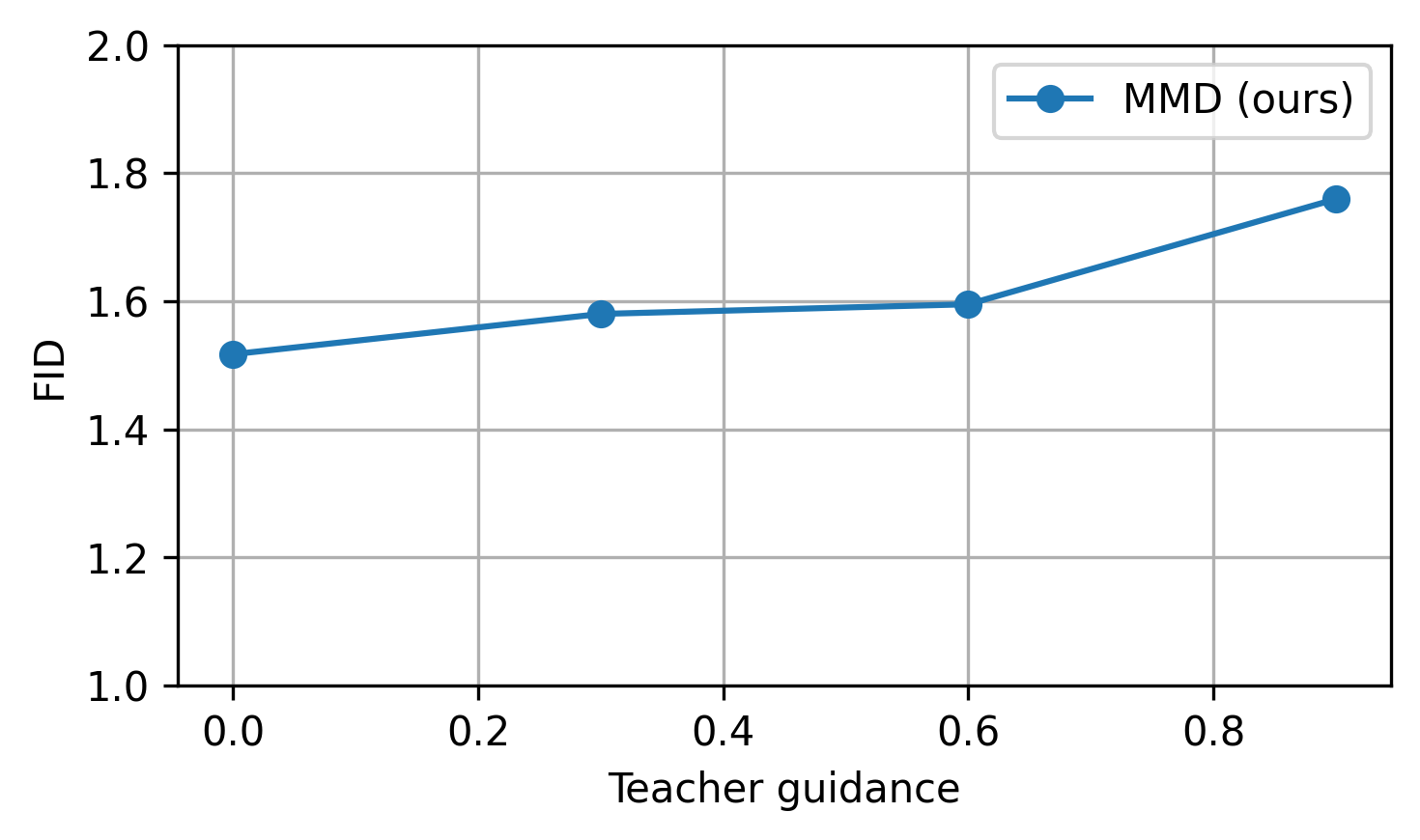}
    \caption{Effect of classifier-free guidance in the teacher model on the FID of a standard MMD student trained in our setup on ImageNet $64\times64$.}
    \label{fig:i64_mmd_guidance}
    \vspace{-10pt}
\end{wrapfigure}

Next, we evaluate the compatibility of Dual-Rate Diffusion with distillation, specifically Moment Matching Distillation (MMD). We chose MMD as a state-of-the-art representative of distribution matching methods whose iterative refinement structure matches Dual-Rate Diffusion most closely: its student samples a stochastic multi-step trajectory, which is exactly the setting in which the Dual-Rate architecture amortizes the context encoder over the intermediate steps. For the student context encoder $E_H$ and the student denoising model $g_\eta$, we use the same parameterization as in the standard diffusion experiments. The teacher model $g_\th$ is our base model, which also initializes the auxiliary model $g_\ph$.

MMD is sensitive to the initialization of the student model, and training a student from scratch leads to instabilities in the early stages. Unlike in regular MMD, the Dual-Rate student cannot be initialized with the teacher model directly, so we pretrain it briefly as a Dual-Rate diffusion model with the context encoder initialized from the teacher (\cref{app:ablations_mmd}).

Following the original MMD setup, we do not use classifier-free guidance for sampling or in the auxiliary model during training. MMD allows guidance in the teacher model during training, but in our experiments it does not improve the student and in some cases destabilizes training, for standard MMD and Dual-Rate MMD alike (\cref{fig:i64_mmd_guidance} shows the former). We therefore train all distilled models with an unguided teacher, which matters for the comparison with published results below.

On ImageNet $64\times64$, Dual-Rate MMD outperforms standard MMD at every step count, reaching a better FID score with a lower computational budget. On ImageNet $128\times128$, Dual-Rate MMD is better at $2$ steps but behind the published MMD results at $4$ and $8$ steps. With only $2$--$8$ denoising steps, there are few intermediate steps over which the cost of the context encoder can be amortized. In the distilled setting the Dual-Rate architecture therefore gains less than in the standard setting, and the gain takes the form of better quality at a comparable budget rather than a lower budget at matched quality.

The published MMD results and ours were also obtained with different teachers (\cref{tab:teachers}). The teacher of \citet{salimans2024multistep} is similar in size to our base model but is used with slight guidance, which makes it roughly one FID point stronger than the unguided base model from which our students are distilled. With $8$ steps, our students nonetheless surpass this stronger teacher at both resolutions, as \citet{salimans2024multistep} also observe for standard MMD students relative to theirs. A direct comparison of the two distillation methods at $128\times128$ would require a shared teacher.

\section{Related Work}
\label{sec:related_work}

\textbf{Accelerating Sampling.} A significant body of research reduces the number of sampling steps in diffusion models, with advanced ODE solvers \citep{lu2022dpm, Karras2022edm, dockhorn2022genie}, distillation methods that compress the sampling process into fewer steps \citep{salimans2022progressive, song2023consistency, yin2023one}, and approaches that learn a direct mapping from noise to data \citep{geng2025mean, deng2026generative}. Our work is orthogonal to these strategies: rather than reducing the number of steps, we lower the computational cost per step, and the two can be combined, as \cref{sec:experiments_mmd} demonstrates for Moment Matching Distillation \citep{salimans2024multistep}.

\textbf{Efficient Architectures and Caching.} The idea of reusing computation across time steps has been explored in several forms. Self-conditioning \citep{chen2022analog} feeds the model's estimate of the clean data from the previous step back as an input to the next step, which improves sample quality without adding parameters; the same heavy network is still evaluated at every step, so the sampling cost is unchanged. Cascaded Diffusion Models \citep{ho2022cascaded} decompose generation into a sequence of models operating at increasing resolutions. Most closely related to our work is Clockwork Diffusion \citep{habibian2023clockwork}, which updates different layers of a single U-Net at different rates by reusing low-resolution feature maps from preceding steps, motivated by the slow evolution of semantic content. DeepCache \citep{ma2024deepcache} similarly reuses high-level U-Net features across steps, and a growing line of work applies caching to Diffusion Transformers, caching tokens by importance scores \citep{lou2024tokencache, zou2024toca} or at random \citep{zou2024duca}, exploiting layer-wise similarity \citep{liu2024smoothcache}, or adapting the caching schedule \citep{kahatapitiya2024adacache}. The caching methods are training-free and accelerate a fixed, already trained network by reusing its internal activations; Clockwork Diffusion trains a light replacement for part of such a network. Dual-Rate Diffusion instead decouples the generative process into two separate networks with independently chosen capacities and trains them jointly. These methods were developed for latent-space U-Nets and Diffusion Transformers rather than for pixel-space UViT models, so we do not compare against them directly; the two approaches are complementary, since caching could be applied on top of a trained Dual-Rate model, to both of its networks.

\section{Limitations and Future Work}
\label{sec:limitations}

Dual-Rate Diffusion increases the training cost per iteration by a factor of $1.4$--$1.5$ in our experiments, since both networks are evaluated at every training step (\cref{sec:experiments_diffusion}). An alternating training scheme for the two networks could reduce this overhead, which we leave for future work.

Dual-Rate Diffusion conditions each denoising step on two states of the noisy sample, $\z_t$ and $\z_\tau$, whose joint distribution must match between training and inference. This restricts training and sampling to stochastic Markovian processes: deterministic samplers degrade the FID (\cref{sec:experiments_diffusion}), and distillation requires rollouts of the student during training (\cref{sec:method_distillation}), which add to its cost.

Several directions remain open. The context encoder is evaluated on a uniform grid in our experiments, but this is a choice of the sampler rather than of the model: \cref{alg:dual_rate_training} samples $\tau$ continuously and only requires an encoder evaluation to be at most $1/K$ ahead of the denoising steps that use it, so a trained model admits non-uniform and adaptive schedules without retraining. In the distilled setting, where times are sampled on the inference grid, such schedules would have to be included in training. Other directions include alternative architectures and conditioning mechanisms for the two networks, applying the construction to latent-space and transformer-based models, and extending it to modalities with pronounced temporal redundancy, such as video and audio, where we expect the gains to be larger.

\section{Conclusion}
\label{sec:conclusion}

We introduced Dual-Rate Diffusion, which accelerates diffusion models by interleaving a heavy context encoder, evaluated at sparse intervals, with a light denoising model that reuses its features at every step. In the standard diffusion setting, it matches or improves on the base model at $2$--$4$ times lower sampling cost. Combined with Moment Matching Distillation, it yields few-step students that sample more cheaply than standard MMD students, improve on them on ImageNet $64\times64$, and at $8$ steps surpass even the stronger teacher of \citet{salimans2024multistep}.

\bibliography{references}

\clearpage
\appendix
\crefalias{section}{appendix}
\crefalias{subsection}{appendix}

\section{Additional results}
\label{app:ablations}

In this section, we present additional results to motivate our design choices and demonstrate the contributions of different components of Dual-Rate Diffusion.

\subsection{Diffusion}
\label{app:ablations_diffusion}

For the context encoder, we use a UViT architecture based on \citet{hoogeboom2025simpler}. We believe this is a reasonable choice, as this architecture is designed to capture long-range dependencies effectively, which is crucial for extracting global structural information from the noisy input.

For the denoising model, we use the same UViT architecture as for the context encoder but with a reduced number of blocks and channels. As the initial configuration, we use the denoising model described in \cref{tab:implementation}. The depth and width ablation of \cref{fig:i64_features_vs_blocks} explores various combinations of channels across the three spatial resolution levels, ranging from $[128, 128, 128]$ to $[256, 256, 512]$, and numbers of transformer blocks at the two transformer levels, ranging from $[1+1, 1]$ to $[3+3, 8]$.

\vspace{6pt}
\noindent\begin{minipage}{\textwidth}
\centering
\captionof{table}{Experimental setup.}
\label{tab:implementation}
\begin{tabular}{lcccc}
\toprule
 & \multicolumn{2}{c}{\textbf{ImageNet $64\times 64$}} & \multicolumn{2}{c}{\textbf{ImageNet $128\times 128$}} \\
\midrule
\quad \textit{Context encoder architecture} & & & & \\
Channels & \multicolumn{2}{c}{$[256, 512, 1024]$} & \multicolumn{2}{c}{$[128, 256, 512, 1024]$} \\
Patch size & \multicolumn{2}{c}{$1\times 1$} & \multicolumn{2}{c}{$2\times 2$} \\
Block types & \multicolumn{2}{c}{[Res, Transf, Transf]} & \multicolumn{2}{c}{[Res, Res, Transf, Transf]} \\
Number of blocks & \multicolumn{2}{c}{$[3+3, 3+3, 16]$} & \multicolumn{2}{c}{$[3+3, 3+3, 3+3, 16]$} \\
Parameters & \multicolumn{2}{c}{$394$M} & \multicolumn{2}{c}{$397$M} \\
GFLOPs per inference & \multicolumn{2}{c}{$108.45$} & \multicolumn{2}{c}{$137.44$} \\
\midrule
\quad \textit{Denoising model architecture} & & & & \\
Channels & \multicolumn{2}{c}{$[256, 256, 512]$} & \multicolumn{2}{c}{$[128, 256, 256, 512]$} \\
Patch size & \multicolumn{2}{c}{$1\times 1$} & \multicolumn{2}{c}{$2\times 2$} \\
Block types & \multicolumn{2}{c}{[Res, Transf, Transf]} & \multicolumn{2}{c}{[Res, Res, Transf, Transf]} \\
Number of blocks & \multicolumn{2}{c}{$[3+3, 3+3, 8]$} & \multicolumn{2}{c}{$[3+3, 3+3, 3+3, 8]$} \\
Parameters & \multicolumn{2}{c}{$88$M} & \multicolumn{2}{c}{$94$M} \\
GFLOPs per inference & \multicolumn{2}{c}{$44.02$} & \multicolumn{2}{c}{$73.01$} \\
\midrule
\quad \textit{Hyperparameters} & & & & \\
Setup & Diffusion & Distillation & Diffusion & Distillation \\
Log-SNR boundaries & $(-12, 12)$ & $(-12, 12)$ & $(-15, 15)$ & $(-15, 15)$ \\
Sigmoid loss bias & $1$ & $1$ & $0$ & $0$ \\
Classifier-free guidance & $1$ & $0$ & $1$ & $0$ \\
Guidance interval & $(1.5, 5)$ & N/A & $(0, 5)$ & N/A \\
Number of TPUs & $16$ & $64$ & $16$ & $64$ \\
Batch size & $1024$ & $1024$ & $1024$ & $512$ \\
Training steps & $10^6$ & $1.5 \times 10^5 $ & $10^6$ & $1.5 \times 10^5 $ \\
Network dropout & $0.1$ & $0.0$ & $0.1$ & $0.0$ \\
Feature dropout & $0.5$ & $0.0$ & $0.5$ & $0.0$ \\
EMA decay & $0.9999$ & $0.999$ & $0.9999$ & $0.999$ \\
Learning rate & $10^{-4}$ & $10^{-6}$ & $10^{-4}$ & $10^{-6}$ \\
Adam betas & $[0.9, 0.99]$ & $[0.0, 0.99]$ & $[0.9, 0.99]$ & $[0.0, 0.99]$ \\
Warmup steps & 0 & $10^3$ & 0 & $10^3$ \\
\bottomrule
\end{tabular}
\end{minipage}
\vspace{6pt}

We also experimented with lighter architectures, such as fully convolutional networks, for the denoising model. While it is possible to reduce the computational cost of model inference by up to 20 times compared to the context encoder, this leads to a significant drop in performance. In our experiments, convolutional denoising models do not reach FID scores below 3 on ImageNet $64\times64$, which is significantly worse than UViT-based architectures. We note that although convolutional blocks are more parameter-efficient than transformer blocks, they remain computationally expensive. A single $3\times 3$ convolutional layer requires $9D^2S^2$ FLOPs, where $D$ is the feature dimension and $S$ is the spatial size. Overall, in our experiments, fully convolutional models do not yield a better performance-compute trade-off than UViT-based architectures.

\subsection{Distillation}
\label{app:ablations_mmd}

\textbf{Initialization of the student.} The Dual-Rate student cannot be initialized with the teacher model directly, because the two have different architectures. We therefore pretrain the student with the standard Dual-Rate objective (\cref{alg:dual_rate_training}), with the context encoder $E_H$ initialized from the teacher and kept frozen, so that only the student denoising model $g_\eta$ is trained. The teacher is a suboptimal feature extractor for the student denoising model, so the pretrained student is weaker than one pretrained from scratch, but this pretraining is considerably cheaper to train, and in our experiments it initializes the distillation equally well.

\cref{alg:dual_rate_mmd_rollout_training} describes the training procedure for Dual-Rate MMD with a full rollout of the student model. It is identical to \cref{alg:dual_rate_mmd_training}, except that $\z_\tau$ is generated by a rollout of the student models $g_\eta$ and $E_H$ starting from pure noise $\z_1 \sim \N(0, \I)$ instead of being sampled from the forward process. Distillation with full rollout therefore does not require data samples.

\begin{algorithm}[h]
\caption{Dual-Rate MMD training algorithm with full rollout of student and alternating optimization of auxiliary denoising model $g_\ph$ and student models $E_H$ and $g_\eta$.}
\label{alg:dual_rate_mmd_rollout_training}
\begin{algorithmic}[1]
\Require Pretrained denoising model $g_\th$, student context encoder $E_H$ and denoising model $g_\eta$, auxiliary denoising model $g_\ph$, number of context encoder evaluations $K$ and denoising steps $k$, loss weight $w(\l_t)$.
\For{$n=0$:$N$}
    \State Sample time of context encoder evaluation $\tau \sim U(\{1/K, \dots, (K-1)/K, 1\})$
    \State Sample step delta $\D_t \sim U(\{0, 1, \dots, k/K - 1\})$ and set $t = \tau - \frac{\D_t}{k}$
    \State Sample time delta $\delta_s \sim u(0, 1/k]$ and set $s = t - \delta_s$
    \State Sample $\z_\tau \sim p_{H,\eta}(\z_\tau)$ with a full rollout of student models $g_\eta$ and $E_H$
    \State Detach $\z_\tau = \sg{\z_\tau}$ and calculate $\ee_\tau = E_H(\z_\tau, \tau)$
    \State Sample $\z_t \sim p_{H,\eta}(\z_t | \z_\tau)$ with a rollout of the denoising model $g_\eta$
    \State Sample $\z_s$ as $\tx = g_\eta(\z_t, t, \ee_\tau), \z_s \sim q(\z_s | \z_t, \tx)$
    \If{$n$ is even}
        \State Minimize $\L = -\frac{\d \l_s}{\d s} e^{\l_s} w(\l_s) \left( \left\| \tx - g_\ph(\z_s, s) \right\|^2 + \left\| g_\th(\z_s, s) - g_\ph(\z_s, s) \right\|^2 \right)$ w.r.t. $\ph$
    \Else
        \State Minimize $\L = -\frac{\d \l_s}{\d s} e^{\l_s} w(\l_s) \tx^\T \sg{g_\ph(\z_s, s) - g_\th(\z_s, s)}$ w.r.t. $\eta$ and $H$
    \EndIf
\EndFor
\end{algorithmic}
\end{algorithm}

For the setup with $K=4$ context encoder evaluations and $k=8$ denoising steps, standard training of Dual-Rate MMD reaches an FID score of $1.61$ on ImageNet $64\times64$, while training with full rollout reaches an FID score of $1.17$ with the same setup.

\section{Implementation Details}
\label{app:implementation}

We provide the implementation details for our experiments on ImageNet $64\times64$ and $128\times128$ in \cref{tab:implementation}. The base model has the architecture of the context encoder, so the Dual-Rate models carry the parameters of both networks in total. For both standard diffusion and distillation, we closely follow the setups from \citet{hoogeboom2025simpler} and \citet{salimans2024multistep}, respectively. We always randomly flip images horizontally with a probability of $0.5$. When using extra data augmentation, we also apply random translation with a probability of $0.4$. We use a cosine noise schedule \citep{nichol2021improved} for all experiments.

For sampling, we use the standard ancestral sampling algorithm adapted for Dual-Rate Diffusion (see~\cref{alg:dual_rate_sampling}). During sampling, we also apply clipping of the $\x$ predictions to the range $[-1, 1]$. For the experiments with standard diffusion, we use classifier-free guidance \citep{ho2022classifier}. To support this, during training, we drop the class conditioning with a probability of $0.1$. In Dual-Rate Diffusion, to obtain conditional and unconditional predictions, we evaluate both the context encoder and the denoising model twice: once with class conditioning and once without. We also use an optimized amount of sampling noise, following \citet{salimans2022progressive}, for standard diffusion, but not for distillation.

We use the same UViT architecture for both the context encoder and the denoising model. To condition the denoising model on two time steps (the current time $t$ and the context encoder evaluation time $\tau$), we use a standard FiLM conditioning mechanism \citep{perez2018film}. To construct a two-time embedding, we calculate the Fourier features for both time steps, project them linearly, and add them together. For the transformer blocks, we use a head dimension of $128$ and, for standard diffusion, dropout. We use the v-parameterization for the denoising model.

We use the Adam optimizer \citep{kingma2014adam} with $\epsilon = 10^{-12}$ and no weight decay. For distillation, we employ a linear learning rate warmup; we do not use learning rate annealing. We use gradient clipping with a maximum norm of $1$. All our experiments are conducted on TPU v6e accelerators.

FID scores are computed on 50k generated samples against the training set (\cref{sec:experiments}). We report the lowest FID score achieved during training, measuring every $20$k iterations for standard diffusion experiments and every $5$k iterations for distillation experiments.

\clearpage
\section{Additional Analysis}
\label{app:analysis}

This section collects the measurements behind the discussions of training cost, deterministic sampling, and teacher quality in \cref{sec:experiments}.

\subsection{Training cost}
\label{app:training_cost}

\begin{table}[h]
\centering
\caption{Training cost of Dual-Rate Diffusion relative to the base model. A training iteration evaluates both networks, so the overhead per iteration is the ratio of the summed per-network FLOPs to those of the base model. Both models are trained for $10^6$ iterations with the same batch size. The matched-compute row reports Dual-Rate Diffusion at the iteration at which its total training compute equals that of the fully trained base model.}
\label{tab:training_cost}
\resizebox{\textwidth}{!}{
\begin{tabular}{lcc}
\toprule
 & \textbf{ImageNet $64\times 64$} & \textbf{ImageNet $128\times 128$} \\
\midrule
GFLOPs per forward pass, context encoder (base model) & $108.45$ & $137.44$ \\
GFLOPs per forward pass, denoising model & $44.02$ & $73.01$ \\
Training cost per iteration, relative to base model & $1.41\times$ & $1.53\times$ \\
\midrule
Iterations at matched training compute & $\approx 7.1 \times 10^5$ & $\approx 6.5 \times 10^5$ \\
FID at matched training compute & $\approx 1.19$ & $\approx 1.57$ \\
FID after $10^6$ iterations & $1.12$ & $1.48$ \\
FID of the base model after $10^6$ iterations & $1.25$ & $1.50$ \\
\bottomrule
\end{tabular}
}
\end{table}

\subsection{Deterministic sampling}
\label{app:ddim}

\begin{table}[h]
\centering
\caption{FID of Dual-Rate Diffusion ($K=16$, $k=512$) with the ancestral sampler used throughout the paper and with a DDIM sampler on the same grid, without sampler-specific tuning.}
\label{tab:ddim}
\begin{tabular}{lcc}
\toprule
Sampler & \textbf{ImageNet $64\times 64$} & \textbf{ImageNet $128\times 128$} \\
\midrule
Ancestral (Markovian) & $1.12$ & $1.48$ \\
DDIM (deterministic) & $1.34$ & $1.80$ \\
\bottomrule
\end{tabular}
\end{table}

\subsection{Teachers in the distillation comparison}
\label{app:teachers}

\begin{table}[h]
\centering
\caption{FID of the teacher models behind the Moment Matching results in \cref{tab:results}. The published MMD results of \citet{salimans2024multistep} use their teacher with guidance $w=0.1$; our distilled models use our base model without guidance, since guidance in the teacher does not improve the student (\cref{fig:i64_mmd_guidance}). The last row gives the FID of the respective $8$-step Dual-Rate MMD student.}
\label{tab:teachers}
\resizebox{\textwidth}{!}{
\begin{tabular}{lcc}
\toprule
 & \textbf{ImageNet $64\times 64$} & \textbf{ImageNet $128\times 128$} \\
\midrule
Teacher of \citet{salimans2024multistep} ($w=0.1$) & $1.42$ & $1.76$ \\
Our base model ($w=1$) & $1.25$ & $1.50$ \\
Our base model ($w=0$), teacher of Dual-Rate MMD & $2.38$ & $2.66$ \\
\midrule
Dual-Rate MMD student ($K=4$, $8$ steps) & $1.17$ & $1.90$ \\
\bottomrule
\end{tabular}
}
\end{table}

\clearpage
\section{Image Samples}
\label{app:samples}

\noindent\begin{minipage}{\textwidth}
    \captionsetup{skip=6pt}
    \centering
    \begin{minipage}[t]{0.48\textwidth}
        \centering
        \includegraphics[width=\linewidth]{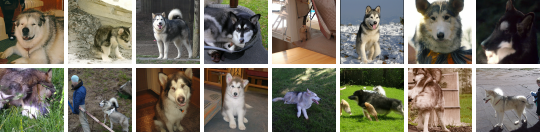}
    \end{minipage}\hfill
    \begin{minipage}[t]{0.48\textwidth}
        \centering
        \includegraphics[width=\linewidth]{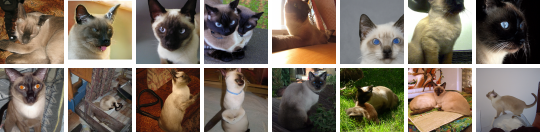}
    \end{minipage}
    \captionof{figure}{Random samples generated by Dual-Rate Diffusion on ImageNet $64\times64$ with $K=16$ context encoder evaluations and $k=512$ denoising steps.}
    \vspace{16pt}
    \begin{minipage}[t]{0.48\textwidth}
        \centering
        \includegraphics[width=\linewidth]{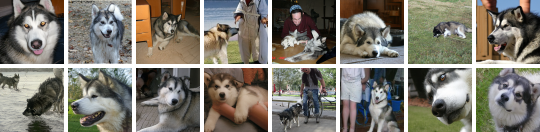}
    \end{minipage}\hfill
    \begin{minipage}[t]{0.48\textwidth}
        \centering
        \includegraphics[width=\linewidth]{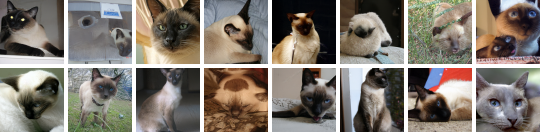}
    \end{minipage}
    \captionof{figure}{Random samples generated by Dual-Rate MMD on ImageNet $64\times64$ with $K=4$ context encoder evaluations and $k=8$ denoising steps.}
    \vspace{16pt}
    \begin{minipage}[t]{0.48\textwidth}
        \centering
        \includegraphics[width=\linewidth]{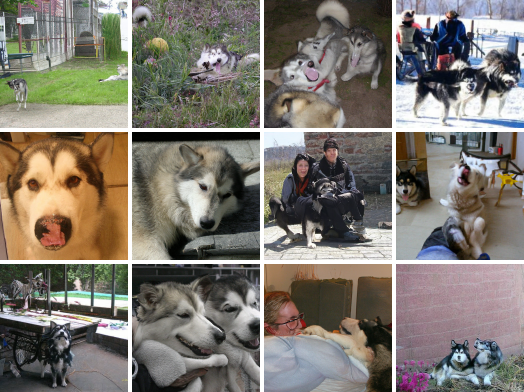}
    \end{minipage}\hfill
    \begin{minipage}[t]{0.48\textwidth}
        \centering
        \includegraphics[width=\linewidth]{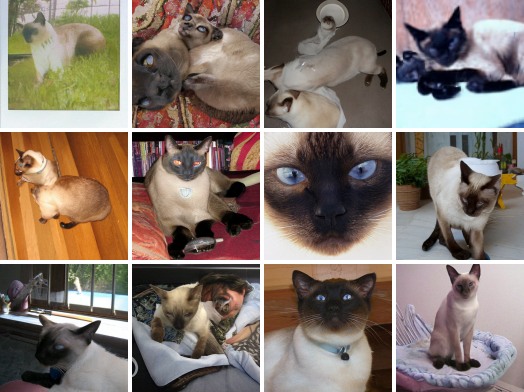}
    \end{minipage}
    \captionof{figure}{Random samples generated by Dual-Rate Diffusion on ImageNet $128\times128$ with $K=16$ context encoder evaluations and $k=512$ denoising steps.}
    \vspace{16pt}
    \begin{minipage}[t]{0.48\textwidth}
        \centering
        \includegraphics[width=\linewidth]{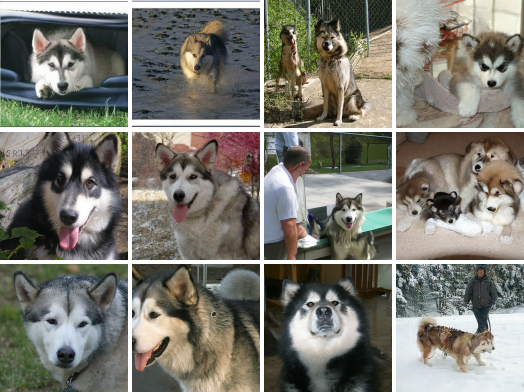}
    \end{minipage}\hfill
    \begin{minipage}[t]{0.48\textwidth}
        \centering
        \includegraphics[width=\linewidth]{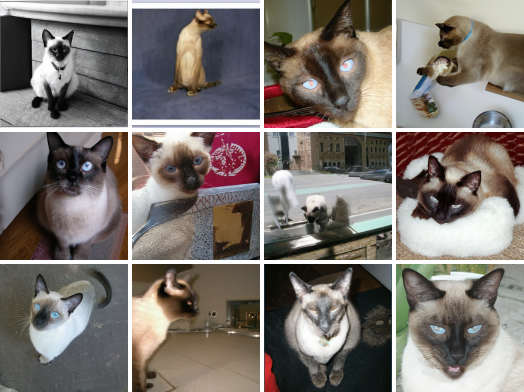}
    \end{minipage}
    \captionof{figure}{Random samples generated by Dual-Rate MMD on ImageNet $128\times128$ with $K=4$ context encoder evaluations and $k=8$ denoising steps.}
\end{minipage}

\end{document}